# An Efficient Algorithm for Computing Interventional Distributions in Latent Variable Causal Models

Ilya Shpitser
Department of Epidemiology
Harvard School of Public Health
ishpitse@hsph.harvard.edu

Thomas S. Richardson
Department of Statistics
University of Washington
tsr@stat.washington.edu

James M. Robins
Department of Epidemiology
Harvard School of Public Health
robins@hsph.harvard.edu

## Abstract

Probabilistic inference in graphical models is the task of computing marginal and conditional densities of interest from a factorized representation of a joint probability distribution. Inference algorithms such as variable elimination and belief propagation take advantage of constraints embedded in this factorization to compute such densities efficiently. In this paper, we propose an algorithm which computes interventional distributions in latent variable causal models represented by acyclic directed mixed graphs (ADMGs). To compute these distributions efficiently, we take advantage of a recursive factorization which generalizes the usual Markov factorization for DAGs and the more recent factorization for ADMGs. Our algorithm can be viewed as a generalization of variable elimination to the mixed graph case. We show our algorithm is exponential in the mixed graph generalization of treewidth.

## 1 Introduction

Establishing cause-effect relationships is fundamental to progress of empirical science. Directed Acyclic Graphs (DAGs) are a popular approach to modeling causation, since they provide an intuitive representation of causal and probabilistic notions. Causal models based on DAGs include structured tree graphs (FFRSTG) [11], non-parametric structural equation models (NPSEM) [6], the minimal causal model (MCM) [12], the so called "agnostic causal models," aka causal Bayesian networks [17], [6], and the "decision theoretic model" [1]. These formalisms represent causal and probabilistic assumptions by means of a directed acyclic graph, with the interpretation of the graph broadly similar, but differing in some details.

One major restriction of these models is that they assume causal sufficiency: causes of at least two variables already in the model must also be in the model. The causal sufficiency assumption is often unreasonable in practical causal inference problems, where partial observability is very common. Approaches to relaxing causal sufficiency include the incorporation of latent variables, or dependent error terms as in the so-called 'semi-Markovian' NPSEMs [6].

Of particular interest in causal inference is the problem of determining causal effects, which are formalized as joint distributions on outcome variables after an idealized experiment called an intervention is performed. A variety of questions in empirical science can be formulated in terms of causal effects. Interventions can sometimes be implemented by randomization, but this frequently cannot be done due to expense or ethical considerations. As a consequence, an important problem is determining in what situations a given causal effect is identifiable from an observational distribution induced by a causal model.

An algorithm for the causal effect identification problem was given in [19],[18],[14], with the completeness of the algorithm established in [15],[14],[3]. This algorithm manipulates the observational distribution and the graph until it either proves the manipulation results in the causal effect of interest or it fails. This process always terminates after a number of recursive calls polynomial in the number of nodes in the graph. However, there is no known efficient scheme for the manipulations of the observational distribution performed by this algorithm. We propose such a scheme here.

This scheme takes advantages of a recursive factorization satisfied by observational distributions of DAG models with latent variables that mirrors the recursive algorithm for identifying causal effects. This recursive factorization implies certain constraints which generalize conditional independence and which we call post-truncation independences. If causally interpretable, these constraints are known as dormant independence constraints [16]. The recursive factorization may also be seen as a refinement of the simpler Markov factorization for ADMGs [8]. Our algorithm can be viewed as a generalization of variable elimination [20] to mixed graphs.

The paper is organized as follows. In Section 2 we define graph theoretic terminology, describe the identification problem for causal effects, and review the ID algorithm which solves it. Section 3 gives a recursive factorization (r-factorization) of distributions with respect to mixed graphs. Section 4 gives a generalized Möbius parameterization of binary r-factorizing models. In Section 5 we describe our efficient algorithm for computing interventional distributions. Section 6 contains the discussion and our conclusions.

## 2 Notation and Basic Concepts

We first introduce the graph-theoretic terminology we need to discuss graphical causal models. In this paper, we will restrict our attention to directed and mixed graphs. A *directed graph* consists of a set of nodes and directed arrows connecting pairs of nodes. A *mixed graph* consists of a set of nodes and directed and/or bidirected arrows ($\leftrightarrow$) connecting pairs of nodes. A *path* is a sequence of distinct nodes where any two adjacent nodes in the sequence are connected by an edge. A *directed path* from a node $X$ to a node $Y$ is a path consisting of directed edges where all edges on the path point away from $X$ and towards $Y$. If a directed edge points from $X$ to $Y$ then $X$ is called a *parent* of $Y$, and $Y$ a *child* of $X$. If $X$ and $Y$ are connected by a bidirected arc, they are called *spouses*. If there is a directed path from $X$ to $Y$ then $X$ is an *ancestor* of $Y$, and $Y$ a *descendant* of $X$; by convention, $X$ is both an ancestor and a descendant of $X$. The sets of parents, children, spouses, ancestors, descendants and non-descendants of a node $X$ are denoted, $Pa(X)$, $Ch(X)$, $Sp(X)$, $An(X)$, $De(X)$ and $Nd(X)$, respectively. A directed *acyclic* graph (DAG) is a directed graph where for any directed path from $X$ to $Y$, $Y$ is not a parent of $X$. An *acyclic directed mixed graph* (ADMG) is a mixed graph which is a DAG if restricted to directed edges.

A *conditional* acyclic directed mixed graph (CADMG) $\mathcal{G}(V, W, E)$ is a mixed graph with a vertex set $V \cup W$, where $V \cap W = \emptyset$, subject to the restriction that for all $w \in W$, $Pa_\mathcal{G}(w) = \emptyset = Sp_\mathcal{G}(w)$. In words, a CADMG is an ADMG with an extra set of nodes $W$ that have no incoming arrows. Intuitively, a CADMG represents a conditional distribution $P(V \mid W)$. For a graph $\mathcal{G}$, and a subset of nodes $S$ in $\mathcal{G}$, we denote by $\mathcal{G}[S]$ the CADMG $\mathcal{G}^*(S, Pa(S)_\mathcal{G} \setminus S, E_S)$, where $E_S$ are those edges in $\mathcal{G}$ that connect nodes in $S$, or directed edges in $\mathcal{G}$ from a node in $Pa(S)_\mathcal{G} \setminus S$ to a node in $S$.

A *district* $S$ in a CADMG $\mathcal{G}$ with vertex set $V \cup W$ is any maximal set of nodes in $V$ that are mutually connected by bidirected paths which are themselves entirely in $S$. For a node $X$, its district in $\mathcal{G}$ is denoted by $D(X)_\mathcal{G}$. The set of all districts in $\mathcal{G}$ forms a partition of nodes in $\mathcal{G}$ and is denoted by $\mathcal{D}(\mathcal{G})$. A set of nodes $S \subseteq V$ in a CADMG $\mathcal{G}$ is called *bi-directed connected* ($\leftrightarrow$-connected for short) if it forms a district in $\mathcal{G}[S]$. For a graph $\mathcal{G}$ over $V$, and vertex set $A$, the *induced sub-graph* $\mathcal{G}_A$ contains only nodes in $A$, and those edges in $\mathcal{G}$ which connect nodes in $A$. A set $A$ is *ancestral* in $\mathcal{G}$ if for all $X \in A$, $An(X)_\mathcal{G} \subseteq A$. The set of all ancestral sets in $\mathcal{G}$ is denoted by $\mathcal{A}(\mathcal{G})$.

A consecutive triple of nodes $W_i, W_j, W_k$ on a path is called a *collider* if the edge between $W_i$ and $W_j$ and the edge between $W_k$ and $W_j$ both have arrowheads pointing to $W_j$. Any other consecutive triple is called a *non-collider*. A path between two nodes $X$ and $Y$ is said to be *blocked* by a set $Z$ if either for some non-collider on the path, the middle node is in $Z$, or for some collider on the path, no descendant of the middle node is in $Z$. For disjoint sets $X, Y, Z$ of nodes in an ADMG we say $X$ is *m-separated* from $Y$ given $Z$ if every path from a node in $X$ to a node in $Y$ is blocked by $Z$. If the ADMG is also a DAG, then we say $X$ is *d-separated* from $Y$ given $Z$. If $X$ is not m-separated (or d-separated) from $Y$ given $Z$, we say $X$ is m-connected (or d-connected) to $Y$ given $Z$. See the graph in Fig. 1 (a) for an illustration of these concepts. In this graph $X_1 \to X_2 \to X_3 \leftarrow X_4$ is a path from $X_1$ to $X_4$; $X_1 \to X_2 \to X_3$ is a directed path from $X_1$ to $X_3$; $X_1$ is a parent of $X_2$, and an ancestor of $X_3$; $X_2 \to X_3 \leftarrow X_4$ is a collider; $X_1$ is m-separated from $X_4$ given $X_5$; $X_1$ is m-connected to $X_4$ given $X_5$ and $X_3$.

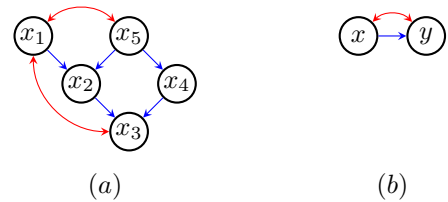

(a)          (b)

Figure 1: (a) An acyclic directed mixed graph. (b) A graph representing some semi-Markovian models where $P(y \mid do(x))$ is not identifiable from $P(v)$.

### 2.1 Causal Models and Causal Effects

A Bayesian network model [5] over variables $V_1, \ldots, V_n$ is a joint probability distribution $P(v_1, \ldots, v_n)$ and a DAG $\mathcal{G}$, such that $P(v_1, \ldots, v_n) = \prod_i P(v_i \mid Pa(V_i)_\mathcal{G} = pa_i)$, where value assignments $pa_i$ are con-sistent with $\{v_1, \ldots, v_n\}$. This decomposition of $P$ into a product is known as a Markov factorization. This factorization is equivalent to the local Markov property which states that each $V_i$ is independent of $\{V_1, \ldots, V_n\} \setminus (De(V_i)_\mathcal{G} \cup Pa(V_i)_\mathcal{G})$ given $Pa(V_i)_\mathcal{G}$, and in turn equivalent to the global Markov property defined by d-separation [5], which states, for any disjoint sets $X, Y, Z$, that if $X$ is d-separated from $Y$ given $Z$ in a graph $\mathcal{G}$, then $X$ is independent of $Y$ given $Z$ in $P(v_1, \ldots, v_n)$. The global Markov property for ADMGs is given by m-separation [9].

Note that we use $V_i$ to refer both to a graph vertex and its associated random variable. In situations where it is important to distinguish vertices and their variables, we will use $V_i$ to refer to the vertex, and $X_{V_i}$ to refer to the variable.

We use $P(y \mid do(x))$ to denote the distribution resulting from an idealized intervention which fixes the value of $X$, and called the *causal effect* of $do(x)$ on $Y$. The causal model associated with a DAG asserts that

$$p(V \setminus X = v^* \mid do(x)) = \prod_{j: V_j \notin X} p(v_j \mid Pa(v_j) = \mathrm{pa}_j);$$

where $v^*$, $x$, $\mathrm{pa}_j$ and the $v_j$ are consistent (in that the same values are assigned to the same variables). This is known as the truncation formula [17],[6], or the g-formula [11].

Note that this definition, and hence our resulting theory, is compatible with any of the causal frameworks currently in the literature (NPSEM, MCM, FFRCISTG, causal BNs, causal IDs).

## 2.2 Latent Projections, Marginal Causal Models, and Causal Effect Identification

Causal models with unobserved variables can be represented by a directed acyclic graph with a subset of nodes labeled as latent. However, another alternative is to represent such models by ADMGs, containing only observed variables, where directed arrows represent "direct causation," and bidirected arrows represent "unspecified causally ancestral latent variables." Every DAG with a set of nodes marked latent can be associated with a 'latent projection', which is an ADMG that preserves path separation relations among the observable variables [7].

For a DAG $\mathcal{G}$ over a variable set $V$, a *latent projection* of $\mathcal{G}$ onto $O \subseteq V$ is an ADMG $\mathcal{G}(O)$, over nodes $O$, where $X_i, X_j \in O$ have an edge $X_i \to X_j$ if there exists a directed path $X_i \to L_1 \to \cdots \to L_k \to X_j$ where $L_1, \ldots L_k \notin O$, and have an edge $X_i \leftrightarrow X_j$ if there exists a trek (that is a collider-free path) from $X_i$ to $X_j$, where all intermediate nodes on the trek are not in $O$, and where the first edge has an arrowhead pointing to $X_i$, and the last edge has an arrowhead pointing to $X_j$.

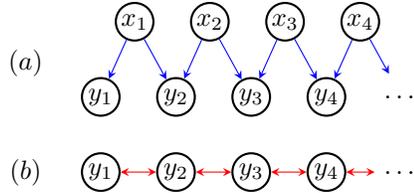

Figure 2: (a) A DAG. (b) A latent projection of the DAG in (a) onto $\{Y_1, Y_2, Y_3, Y_4, \ldots\}$.

Latent projections are a convenient way to represent latent variable models because they simplify the representation of multiple paths of latents while preserving many useful constraints over observable variables; see Figure 2 for an example.

Consider an ADMG $\mathcal{G}$ induced by some latent variable causal model, and disjoint sets $X, Y$ of nodes in this $\mathcal{G}$. Is it possible to characterize situations when the causal effect of $do(x)$ on $Y$ is identifiable in any model inducing $\mathcal{G}$? It turns out that the answer is yes, and there is an explicit, closed-form algorithm for deriving the functional for such a causal effect whenever it is identifiable. This algorithm was first developed in [19], for causal Bayesian networks and semi-Markovian models. A version of this algorithm, called ID, which appears in [15] is shown in Fig. 3.

The ID algorithm is recursive, using simplification methods, based on districts and ancestral sets. First, the algorithm eliminates any irrelevant variables by intervening on them with arbitrary values (line 3). Second, the algorithm splits the problem into subproblems based on districts (line 4). Third, the algorithm marginalizes out some variables such that the margin that is left is an ancestral set in the graph resulting from interventions (line 2). And finally, the algorithm "truncates out" some variables in situations where the DAG truncation formula applies to a particular intervention (lines 6 and 7). When the algorithm fails (line 5), it returns a witness for this failure, called a 'hedge':

**Definition 1 (c-forest)** *Let $R$ be a subset of nodes in an ADMG $\mathcal{G}$. Then a set $F$ is called an $R$-rooted C-forest if $R \subseteq F$, $F$ is a $\leftrightarrow$-connected set, and every node in $F$ has a directed path to a node in $R$ with every element on the path also in $F$.*

**Definition 2 (hedge)** *Let $X, Y$ be disjoint node sets in an ADMG $\mathcal{G}$. Then two $R$-rooted C-forests $F, F'$*

function **ID**$(y, x, P(V), \mathcal{G}(V))$
INPUT: $x, y$ value assignments, $P(V)$ a probability distribution over $V$, $\mathcal{G}(V)$ an ADMG over nodes $V$.
OUTPUT: Expression for $P_x(y)$ in terms of $P(V)$ or a hedge $(F, F')$ witness for non-identification.

1. if $x = \emptyset$, return $\sum_{v \setminus y} P(V)$.

2. if $V \setminus An(Y)_\mathcal{G} \neq \emptyset$, return
   **ID**$(y, x_{An(Y)_\mathcal{G}}, \sum_{v \setminus An(Y)_\mathcal{G}} P(V), \mathcal{G}(An(Y)))$.

3. let $W = (V \setminus X) \setminus An(Y)_{\mathcal{G}[V \setminus X]}$.
   if $W \neq \emptyset$, return **ID**$(y, x \cup w, P(V), \mathcal{G}(V))$.

4. if $\mathcal{D}(\mathcal{G}[V \setminus X]) = \{S_1, \ldots, S_k\}$,
   return $\sum_{v \setminus (y \cup x)} \prod_i$ **ID**$(s_i, v \setminus s_i, P(V), \mathcal{G}(V))$.

   if $\mathcal{D}(\mathcal{G}[V \setminus X]) = \{S\}$,

   5. if $\mathcal{D}(\mathcal{G}) = \{V\}$, stop with $(V, V \cap S)$.
   6. if $S \in \mathcal{D}(\mathcal{G})$, return $\sum_{s \setminus y} \prod_{\{i | V_i \in S\}} P(V_i | \overline{V_i})$.
   7. if $(\exists S')\ S \subset S' \in \mathcal{D}(\mathcal{G})$ return **ID**$(y, x \cap S', \prod_{\{i|V_i \in S'\}} P(V_i | \overline{V_i} \cap S', \overline{v_i} \setminus S'), \mathcal{G}_{S'})$.

Figure 3: A complete identification algorithm. $\overline{V_i}$ is the set of nodes preceding $V_i$ in some topological ordering $\pi$ in $\mathcal{G}$.

form a hedge for $(X, Y)$ in $\mathcal{G}$ if $F \subset F'$, $R \subset An(Y)_{\mathcal{G}[V \setminus X]}$, and $X \subset F' \setminus F$.

Hedges characterize non-identification of causal effects due to the following theorem [15].

**Theorem 1** $P(y \mid do(x))$ is identifiable in any semi-Markovian model inducing an ADMG $\mathcal{G}$ if and only if there is no hedge for any $(X', Y')$ in $\mathcal{G}$, where $X' \subseteq X, Y' \subseteq Y$, and $X'$ is non-empty.

The simplest graph where causal effect identification is not possible is shown in Fig. 1(b). In this graph, the hedge witnessing non-identifiability of $P(y \mid do(x))$ is the two sets $\{Y\}$, and $\{X, Y\}$. Mapping to the definition, $R = \{Y\}, F = \{Y\}, F' = \{X, Y\}$.

### 2.3 Evaluating Causal Effect Functionals

Consider the graph shown in Fig. 7(a), where we wish to use **ID** to evaluate the causal effect of $do(x_3)$ on $X_5$, or $P(x_5 \mid do(x_3))$. Applying **ID** to this task yields the following identity:

$P(x_5 \mid do(x_3)) =$
$= \sum_{x'_4} \left( \sum_{x'_2} P(x_4 \mid x_3, x'_2, \tilde{x}_1) P(x'_2 \mid \tilde{x}_1) \right) \times$
$\left( \sum_{x'_1, x'_3} P(x_5 \mid x'_4, x'_3, \tilde{x}_2, x'_1) P(x'_3 \mid \tilde{x}_2, x'_1) P(x'_1) \right)$

where $\tilde{x}_2$ and $\tilde{x}_1$ are arbitrary values. This functional involves the term $P(x_5 \mid x_4, x'_3, \tilde{x}_2, x'_1)$, which in binary models corresponds to 32 numbers. As mixed graphs increase in size, the size of conditional probability tables necessary to evaluate causal effect functionals quickly becomes intractable. In DAG models, inference algorithms based on the factorization systematically exploit conditional independence constraints implied by this factorization to make inference computationally tractable. To extend this approach to mixed graphs, we introduce a recursive factorization.

## 3 Recursive Factorization of ADMGs

The recursive factorization is defined in terms of special sets of nodes called 'intrinsic' sets, which are closely related to certain identifiable causal effects, though the notion of an intrinsic set is purely graphical.

**Definition 3** Let $\mathcal{G}$ be an ADMG. Then a set of nodes $S$ in $\mathcal{G}$ is intrinsic if it is $\leftrightarrow$-connected and there does not exist a hedge for $(Pa(S) \setminus S, S)$ in $\mathcal{G}$.

An alternative but equivalent definition of intrinsic sets exists, which does not involve concepts from identification.

**Definition 4** Let $S$ be a $\leftrightarrow$-connected set in an ADMG $\mathcal{G}$. Define the following operations on subgraphs of $\mathcal{G}$:

$$\begin{aligned} \alpha_S : & \quad \mathcal{G} \mapsto \mathcal{G}_{An_\mathcal{G}(S)}, \\ \delta_S : & \quad \mathcal{G} \mapsto \mathcal{G}_{D_\mathcal{G}(S)}, \\ \gamma_S : & \quad \mathcal{G} \mapsto \alpha_S(\delta_S(\mathcal{G})), \\ \gamma_S^{(k)} : & \quad \mathcal{G} \mapsto \underbrace{\gamma_S(\ldots \gamma_S(\mathcal{G}) \ldots)}_{k\text{-times}}. \end{aligned}$$

Then $S$ is intrinsic in $\mathcal{G}$ if $(\exists k > 0)$, $\mathcal{G}_S = \gamma_S^{(k)}(\mathcal{G})$.

We will denote the set of all intrinsic sets in $\mathcal{G}$ by $\mathcal{I}(\mathcal{G})$. Every $\leftrightarrow$-connected set $S$ in $\mathcal{G}$ is contained in a unique smallest member of $\mathcal{I}(\mathcal{G})$, called the intrinsic closure of $S$.

Our factorization is recursive, and thus is defined on CADMGs, in which the set $W$ may be thought of as representing 'pre-existing interventions'.

**Definition 5 (r-factorization)** Let $\mathcal{G}(V, W, E)$ be a CADMG and $\mathfrak{F}_\mathcal{G} \equiv \{f_C(X_C \mid X_{Pa(C) \setminus C}), C \in \mathcal{I}(\mathcal{G})\}$ be a collection of densities. A distribution $p(X_V \mid X_W)$ recursively factorizes (r-factorizes) according to $\mathcal{G}(V, W, E)$ and $\mathfrak{F}_\mathcal{G}$ if for every subset $A \subseteq V$ such that $an_\mathcal{G}(A) \cap V = A$ the following hold:

(i) $p(X_A \mid X_W) = \prod_{D \in \mathcal{D}(\mathcal{G}_{A \cup W})} f_D(X_D \mid X_{Pa(D) \setminus D})$

(ii) if $|\mathcal{D}(\mathcal{G}_{A \cup W})| > 1$ then for every district $D \in \mathcal{D}(\mathcal{G}_{A \cup W})$, $f_D(X_D \mid X_{Pa(D) \setminus D})$ r-factorizes according to $(\mathcal{G}_{A \cup W})[D]$ and the set of densities $\{f_C \in \mathfrak{F}_\mathcal{G}, C \in \mathcal{I}((\mathcal{G}_{A \cup W})[D])\}$.

Note that this factorization can be viewed as a refinement of the factorization in [8] which only represented conditional independence constraints (corresponding to m-separations), whereas the criterion here also represents conditional independences after division by a conditional density. If such a division can be causally interpreted as an intervention, such independences are called dormant in [16].

We now show that the observed marginals of causal DAG models r-factorize.

**Theorem 2** *Let $P(O)$ be the observed marginal of a causal DAG, with latent projection given by an ADMG $\mathcal{G}$. Then there exists $\mathfrak{F}_\mathcal{G}$ such that $P(O)$ r-factorizes according to $\mathcal{G}$ and $\mathfrak{F}_\mathcal{G}$.*

*Proof:* Let $\mathfrak{F}_\mathcal{G} = \{P(c \mid do(Pa(C) \setminus C = pa)) \mid C \in \mathcal{I}(\mathcal{G}), pa$ assignments to $Pa(C) \setminus C\}$. The theorem follows by soundness of the **ID** algorithm. □

This theorem implies that our results apply to any causal DAG model for which $P(v \mid do(x))$ is given by the g-formula.

## 4 Parameterization of Binary R-factorizing Models

We now give a parameterization of r-factorizing binary models. The approach generalizes in a straightforward way to finite discrete state spaces.

### 4.1 Heads of intrinsic sets

To parameterize our models, we will associate a set of parameters with each intrinsic set, just as binary DAG models associate parameters of the form $P(X_i = 0 \mid Pa(X_i) = pa_i)$ with each node $X_i$. We first define sets of nodes, called 'heads' and 'tails' related to a given set $C \in \mathcal{I}(\mathcal{G})$ that will serve the role of $X_i$ and $Pa(X_i)$, respectively, in the DAG case.

**Definition 6** *For an intrinsic set $C \in \mathcal{I}(\mathcal{G})$ of a CADMG $\mathcal{G}$, define the* recursive head (rh) *as: $rh(C) \equiv \{x \mid x \in C; ch_{\mathcal{G}_C}(x) = \emptyset\}$. Let $\mathcal{RH}(\mathcal{G}) \equiv \{rh(C) \mid C \in \mathcal{I}(\mathcal{G})\}$.*

**Definition 7** *The* tail *associated with a recursive head $H$ of an intrinsic set $C$ in a CADMG $\mathcal{G}$ is given by: $tail(H) \equiv (C \setminus H) \cup pa_\mathcal{G}(C)$*

### 4.2 Recursive head partition of arbitrary sets

We now show how to partition an arbitrary subset of $V$ in a CADMG $\mathcal{G}(V, W, E)$ into elements of $\mathcal{RH}(\mathcal{G})$.

**Definition 8** *Let $\prec_{\mathcal{I}(\mathcal{G})}$ be a partial order on heads of intrinsic sets of $\mathcal{G}$ such that $H_1 \prec_{\mathcal{I}(\mathcal{G})} H_2$ if $H_1 = rh(C_1), H_2 = rh(C_2), C_1, C_2 \in \mathcal{I}(\mathcal{G}), C_1 \subseteq C_2$*

For a set of heads $\mathcal{H}$, let $max_{\prec_{\mathcal{I}(\mathcal{G})}}(\mathcal{H})$ be the subset of $\mathcal{H}$ containing heads maximal in $\mathcal{H}$ under $\prec_{\mathcal{I}(\mathcal{G})}$.

**Definition 9** *For any $B \subseteq V$ of nodes in a CADMG $\mathcal{G}(V, W, E)$, define the following functions:*

$$\Upsilon_\mathcal{G}(B) \equiv max_{\prec_{\mathcal{I}(\mathcal{G})}}(\mathcal{RH}(\mathcal{G}) \cap \mathcal{P}(B))$$

$$\rho_\mathcal{G}(B) \equiv B \setminus \bigcup_{H \in \Upsilon_\mathcal{G}(B)} H, \quad \rho_\mathcal{G}^{(k)}(B) \equiv \underbrace{\rho_\mathcal{G}(\cdots \rho_\mathcal{G}(B) \cdots)}_{k\text{-times}}$$

*where $\mathcal{P}(B)$ is the power set of $B$, and $\rho_\mathcal{G}^{(0)}(B) \equiv B$. We partition $B$ into recursive heads of $\mathcal{G}$ as follows:*

$$[\![B]\!]_\mathcal{G} \equiv \bigcup_{k \geq 0} \Upsilon_\mathcal{G}\left(\rho_\mathcal{G}^{(k)}(B)\right)$$

### 4.3 Binary Parameterization

Multivariate binary distributions which r-factorize with respect to a CADMG $\mathcal{G}$ may be parameterized by the following parameters:

**Definition 10** *The* binary parameters *associated with CADMG $\mathcal{G}$ are a set of functions $\mathfrak{Q}_\mathcal{G}$:*

$$\left\{q_H(x_{tail(H)}) \mid H \in \mathcal{RH}(\mathcal{G}), q_H : \{0,1\}^{|tail(H)|} \to [0,1]\right\}$$

Intuitively the functions $q_H(x_{tail(H)})$ should be thought of as giving the value $f_C(X_H = 0 \mid X_{tail(H)} = x_{tail(H)})$, where $H = rh_\mathcal{G}(C)$. We emphasize that the density $f_C$ is not necessarily a standard conditional density in $P$. Instead in any causal model where the g-formula holds, $f_C(X_H = 0 \mid X_{tail(H)} = x_{tail(H)})$ can be interpreted as a conditional interventional distribution $P(X_H = 0 \mid X_{C \setminus H} = x_{C \setminus H}, do(X_{Pa(C) \setminus C} = x_{Pa(C) \setminus C}))$, where $H = rh(C)$.

**Definition 11** *Let $\nu : V \cup W \mapsto \{0, 1\}$ be an assignment of values to the variables indexed by $V \cup W$. Define $\nu(T)$ to be the values assigned to variables indexed by a subset $T \subseteq V \cup W$. Let $\nu^{-1}(0) = \{v \mid v \in V, \nu(v) = 0\}$.*

function **EID**$(y, x, \mathfrak{Q}(V), \mathcal{G}(V))$
INPUT: $x, y$ value assignments, $\mathfrak{Q}(V)$ a set of parameters representing a distribution $P(v)$ which r-factorizes with respect to $\mathcal{G}$, $\mathcal{G}$ an ADMG.
OUTPUT: A set of $q$ parameters representing $P_x(y)$, or **FAIL**.

1. Let $V^* = An(Y)_{\mathcal{G}[V \setminus X]} \setminus X$, $\mathfrak{Q}^* = \mathfrak{Q}(V^* \mid x)$, $\mathcal{G}^* = \mathcal{G}[V^*]$.

2. If $\mathcal{I}(\mathcal{G}[V^*]) \not\subseteq \mathcal{I}(\mathcal{G}(V))$, return **FAIL**.

3. Pick an elimination order $V_1, \ldots, V_k$ for $V^* \setminus Y$.

4. For each $i = 1, \ldots, k$, do:
   $(\mathfrak{Q}^*, \mathcal{G}^*) \leftarrow$ **sum-one**$(V_i, \mathfrak{Q}^*, \mathcal{G}^*)$.

5. return $\mathfrak{Q}^*$

Figure 4: An efficient algorithm for computing marginal interventional distributions.

A distribution $P(X_V \mid X_W)$ is said to be *parameterized by* the set $\mathfrak{Q}_\mathcal{G}$, for CADMG $\mathcal{G}$ if:

$$p(X_V = \nu(V) \mid X_W = \nu(W)) = \sum_{B \,:\, \nu^{-1}(0) \cap V \subseteq B \subseteq V} (-1)^{|B \setminus \nu^{-1}(0)|}$$
$$\times \prod_{H \in [\![B]\!]_\mathcal{G}} q_H(X_{tail(H)} = \nu(tail(H))),$$

*where the empty product is defined to be* 1.

Note that this parameterization maps $q_H$ parameters to probabilities in a CADMG via an inverse Möbius transform. Note also that this parameterization generalizes both the standard Markov parameterization of DAGs in terms of parameters of the form $p(X_i = 0 \mid Pa(x_i) = pa_i)$, and the Möbius parameterization of bidirected graph models given in [2]. We will denote the generalized Möbius transform which maps a set $\mathfrak{Q}(V) = \{q_H(tail(H)) \in \mathfrak{Q}_\mathcal{G} \mid H \subseteq V\}$, to a distribution $P(X_V = \nu(V) \mid X_W = \nu(W))$ as **GMT**$(\mathcal{G}(V), \mathfrak{Q}(V), \nu)$. In a companion paper [10] we show that a multivariate binary distribution $P$ is parameterized by $\mathfrak{Q}(V)$ if and only if $P$ r-factorizes according to $\mathcal{G}$. We will call the set of $q$ parameters representing a (possibly conditional) distribution $P$ the $q$ *representation of* $P$.

## 5 An Efficient Algorithm for Computing Interventional Marginals

In this section, we describe our inference algorithm for computing causal effect marginals. Due to Theorem 2, our algorithm applies to any graphical causal model where the g-formula holds. The algorithm is shown in Figs. 4 and 5.

The algorithm returns a $q$ representation of an identifiable query $P(y \mid do(x))$ in two stages. In the first stage, the algorithm computes a $q$ representation of the distribution $P(An(y)_{\mathcal{G}([V \setminus X])} \setminus X \mid do(x))$ from a $q$ representation of $P(v_1, \ldots, v_n)$, denoted by $\mathfrak{Q}(V)$. As we will show in the next section, a representation of this marginal interventional can be computed in one step, by restricting the set $\mathfrak{Q}(V)$ to those parameters $q_H(tail(H))$ with $H \subseteq \mathcal{RH}(\mathcal{G}) \cap V^*$, where $V^* = An(Y)_{\mathcal{G}[V \setminus X]} \setminus X$, and with tail assignments consistent with $x$. In the algorithm, this restriction is denoted by $\mathfrak{Q}(V^* \mid x)$.

The second stage marginalizes out variables not in $Y$ from $P(An(y)_{\mathcal{G}([V \setminus X])} \setminus X \mid do(x))$ using a particular elimination order. Marginalizing a single variable $X$ is done by converting from a $q$ representation to a standard probability representation, marginalizing $X$ from the table, and converting back to a $q$ representation.

In the remainder of the section, we are going to show that the algorithm is sound (i.e.. performs probabilistic calculations correctly), and then analyze its time and space complexity.

### 5.1 Soundness

To show soundness of **EID**, we will first show that the set of unintervened nodes in the post-intervention world considered by **ID** before the final marginalization step in lines 1, 4, or 6 consists precisely of nodes in $An(Y)_{\mathcal{G}[V \setminus X]} \setminus X$. Next, we show that the distribution corresponding to that post-intervention world can be computed by restricting the set of parameters onto the set of nodes in that world. Finally, we show the soundness of the last marginalization step by showing **sum-one** is sound.

**Lemma 1** *For a given query* $(y, x, P(v), \mathcal{G}(V))$, *line 4 of* **ID** *is invoked at most once.*

*Proof:* If line 4 is invoked, all subproblems contain outcome sets which form a $\leftrightarrow$-connected set. Since no recursive call of **ID** considers subsets of such outcome sets, line 4 is never invoked on any subproblem. □

For any recursive call **ID**, let $V^*$ be the set of nodes in the ADMG $\mathcal{G}^*$ passed to that recursive call's scope.

**Lemma 2** *If* **ID** *is invoked with* $(y, x, P(v), \mathcal{G}(V))$, *and terminates successfully, then in the terminating line* **ID** *evaluates an expression corresponding to* $\sum_{v^* \setminus y} P(v^* \mid do(x))$, *where* $An(Y)_{\mathcal{G}[V \setminus X]} \setminus X \subseteq V^* \setminus X$ *and* $An(Y)_{\mathcal{G}[V \setminus X]} \setminus X \in \mathcal{A}(\mathcal{G}^*[V^* \setminus X])$.

function **sum-one**$(X, \mathfrak{Q}(V), \mathcal{G}(V))$
INPUT: $X$ a single variable, $\mathfrak{Q}(V)$ a set of parameters representing a distribution $P(v)$ which r-factorizes with respect to $\mathcal{G}(V)$, $\mathcal{G}(V)$ an ADMG.
OUTPUT: A set of $q$ parameters representing $\sum_x P(v)$, and a latent projection $\mathcal{G}(V \setminus \{X\})$.

1. Compute $\mathcal{G}^* = \mathcal{G}(V \setminus \{X\})$ from $\mathcal{G}(V)$.
2. Compute $\mathcal{I}_x = \mathcal{I}(\mathcal{G}^*)$.
3. Let $\mathfrak{Q}_{sum} = \{\}$. For each $I \in \mathcal{I}_x$ where $X \in Pa(I)_{\mathcal{G}(V)}$, do
   3a. Let $H = rh(I)$, $T = (I \cup Pa(I)_{\mathcal{G}^*}) \setminus rh(I)$.
   3b. For all assignments $\nu$ to $I \cup \{X\}$ where $\nu(H) = 0, \nu(T) = t$, order $\nu$ in ascending order of cardinality of 1 assignments in $\nu$, evaluate in order: $q_h(T = \nu(T)) = \frac{\sum_x \prod_{D \in \mathcal{D}(\mathcal{G}(V)[I \cup \{X\}])} \mathbf{GMT}(\mathcal{G}(V)[D], \mathfrak{Q}(D), \nu)}{\sum_x \prod_{D \in \mathcal{D}(\mathcal{G}(V)[(I \cup \{X\}) \setminus rh(I)])} \mathbf{GMT}(\mathcal{G}(V)[D], \mathfrak{Q}(D), \nu)}$.
   3c. $\mathfrak{Q}_{sum} \leftarrow \mathfrak{Q}_{sum} \cup \{q_h(T = \nu(T))\}$.
4. $\mathfrak{Q}^* \leftarrow \{q_{rh(I)}((I \cup Pa(I)) \setminus rh(I)) \in \mathfrak{Q}(V) \mid I \in \mathcal{I}_x, X \notin Pa(I)_{\mathcal{G}(V)}\} \cup \mathfrak{Q}_{sum}$.
5. return $(\mathfrak{Q}^*, \mathcal{G}^*)$.

Figure 5: An algorithm for computing $q$ parameters of an r-factorizing model obtained from marginalizing a single variable $X$ from the input model.

*Proof:* Successful termination occurs in lines 1, 4, and 6. $An(Y)_{\mathcal{G}[V \setminus X]} \setminus X \subseteq V^* \setminus X$ is established by function call induction. Certainly $An(Y)_{\mathcal{G}[V \setminus X]} \cap \subseteq V \setminus X$, where $V$ is the set of all nodes in the outermost call. The only lines which remove nodes are 2 and 7. We have two cases. If line 4 is never called, then in line 2, by inductive hypothesis, $\mathcal{G}^*$ is a supergraph of $\mathcal{G}[An(Y)_{\mathcal{G}[V \setminus X]} \setminus X]$. This means $An(Y)_{\mathcal{G}[V \setminus X]} \setminus X \subseteq An(Y)_{\mathcal{G}^*}$. In line 7, only a subset of nodes in $X$, say $X^*$, is removed. By inductive hypothesis, $An(Y)_{\mathcal{G}[V \setminus X]} \setminus X \subseteq V \setminus X^*$. If line 4 is called, it is only once due to Lemma 1. We have $An(Y)_{\mathcal{G}[V \setminus X]} \setminus X \subseteq V^* \setminus X$, by induction. Finally, $\mathcal{G}^*[V^* \setminus X]$ is a subgraph of $\mathcal{G}[V \setminus X]$, so any set ancestral in $\mathcal{G}[V \setminus X]$ is also ancestral in $\mathcal{G}^*[V^* \setminus X]$. But we know $An(Y)_{\mathcal{G}[V \setminus X]}$ is ancestral in $\mathcal{G}[V \setminus X]$. □

**Lemma 3** *In a CADMG $\mathcal{G}$, for any $A \subseteq (V \cup W)$ such that $An_{\mathcal{G}}(A) = A$, $\mathcal{I}(\mathcal{G}_A) = \mathcal{I}(\mathcal{G}) \cap \mathcal{P}(A \cap V)$, where $\mathcal{P}(\cdot)$ denotes the powerset.*

**Lemma 4** *In a CADMG $\mathcal{G}$, for any district $D$, $\mathcal{I}(\mathcal{G}_D) = \mathcal{I}(\mathcal{G}) \cap \mathcal{P}(D)$.*

Finally, we are ready to prove that the first stage of **EID** is sound.

**Lemma 5** *If $P(y \mid do(x))$ is identifiable in $\mathcal{G}(V)$, and $P(v)$ r-factorizes according to $\mathcal{G}(V)$, then $P(An(y)_{\mathcal{G}[V \setminus X]} \setminus X \mid do(x))$ is identifiable in $\mathcal{G}(V)$, and is parameterized by $\mathfrak{Q}(An(Y)_{\mathcal{G}[V \setminus X]} \setminus X \mid x)$ in $\mathcal{G}[An(Y)_{\mathcal{G}[V \setminus X]} \setminus X]$.*

*Proof:* The distribution operations in **ID** corresponding to lines 2 and 7 are ancestral marginalizations and interventions on all nodes outside a districts. Since the $q$ parameters are fully determined by the distributions $P(X_C \mid X_{tail(C)})$, the previous three lemmas imply our result. □

To prove the soundness of stage two, we have two cases. If $I \in \mathcal{I}(\mathcal{G}(V \setminus \{X\}))$ and $X \notin Pa(I)_{\mathcal{G}(V)}$, then $I \in \mathcal{I}(\mathcal{G}(V))$. Further, by Lemma 3, the $q$ parameters associated with $I$ remain unchanged in $\mathcal{G}(V \setminus \{X\})$. Otherwise, we need the following lemma.

**Lemma 6** *If $I \in \mathcal{I}(\mathcal{G}(V \setminus \{X\}))$ and $X \in Pa(I)_{\mathcal{G}(V)}$, then $\forall D \in \mathcal{D}(\mathcal{G}[I \cup \{X\}]), D \in \mathcal{I}(\mathcal{G}(V))$.*

*Proof:* By construction, each such $D$ is ↔-connected. We want to show the distribution $P(i \cup \{x\} \mid do(Pa(I \cup \{X\})_{\mathcal{G}(V)} \setminus (I \cup \{X\}) = pa_i))$ is identifiable in $\mathcal{G}(V)$. Once we do so, our conclusion will follow, since for each $D$, $P(d \mid do(Pa(D)_{\mathcal{G}} \setminus D = pa_d))$ is identifiable from $P(I \cup \{X\} \mid do(Pa(I \cup \{X\})_{\mathcal{G}(V)} \setminus (I \cup \{X\}) = pa_i))$.

Assume for contradiction $P(i \cup \{x\} \mid do(Pa(I \cup \{X\})_{\mathcal{G}(V)} \setminus (I \cup \{X\}) = pa_i))$ is not identifiable in $\mathcal{G}(V)$. Then there exists a hedge for $(Pa(I \cup \{X\}) \setminus (I \cup \{X\}), I \cup \{X\})$ in $\mathcal{G}(V)$. Let $R$ be the subset of $I \cup \{X\}$ with no children in $\mathcal{G}[I \cup \{X\}]$. By definition, there exists a hedge for $(Pa(I \cup \{X\}) \setminus (I \cup \{X\}), R)$. If any vertex $Z$ in $I$ has $X$ as a child in $\mathcal{G}[I \cup \{X\}]$, then it will have a child in $\mathcal{G}[I]$, hence gets nothing added to $R$ in $\mathcal{G}(V \setminus \{X\})$. Note also that the sets $Pa(I \cup \{X\})_{\mathcal{G}(V)} \setminus (I \cup \{X\})$, and $Pa(I)_{\mathcal{G}(V \setminus \{X\})} \setminus I$ are the same. Finally, since $X$ contains a child $C$ in $I$, any C-forest $F$ in $\mathcal{G}(V)$ either remain in $\mathcal{G}(V \setminus \{X\})$, if $F$ does not intersect $X$, or an isomorphic C-forest $F'$ exists in $\mathcal{G}(V \setminus \{X\})$, where $X$ is replaced by $C$. This implies a contradiction, since we just showed the existence of a hedge for $(Pa(I)_{\mathcal{G}(V \setminus \{X\})} \setminus I, I)$, but we assumed $I \in \mathcal{I}(\mathcal{G}(V \setminus \{X\}))$. □

This lemma coupled with the soundness of the **GMT** mapping in the previous section establishes soundness of **sum-one**.

### 5.2 Mixed Graph Binary Width

We now introduce a certain measure of graph "width," which is derived from the dimension of the model. We will show that the computational complexity of **EID** is exponential in this measure. This width can be viewed as a tally of those parameters which we compute by invoking **GMT**, for a particular elimination order.

**Definition 12** *For an ADMG $\mathcal{G}(V)$ and a node $X \in V$, let $\mathcal{H}_x = \{rh(I) \mid I \in \mathcal{I}(\mathcal{G}(V \setminus \{X\})), X \in Pa(I)_{\mathcal{G}(V)}\}$. The* binary width *of $\mathcal{G}(V)$ with respect to $X$, or $bw(\mathcal{G}(V), X)$ is equal to $\log_2 \sum_{H \in \mathcal{H}_x} 2^{|tail(H)|}$.*

**Definition 13** *For an elimination order $\pi$ indexing nodes $Z_1, \ldots, Z_k$ in a set $Z \subset V$ in $\mathcal{G}(V)$, the binary width of $\mathcal{G}(V)$ using $\pi$ with respect to $Z$ is $bw_\pi(\mathcal{G}(V), Z) = \max_i bw(\mathcal{G}(V \setminus Z, Z_i, \ldots, Z_k), Z_i)$. The binary width of $\mathcal{G}(V)$ with respect to a set $Z \subseteq V$, written as $bw(\mathcal{G}(V), Z)$ is the smallest binary width across all elimination orderings of nodes in $Z$.*

Note that considering an interventional query $P(y \mid do(x))$ will often reduce the width, since evaluating such a query entails considering the mutilated graph $\mathcal{G}[An(Y)_{\mathcal{G}[V \setminus X]} \setminus X]$ which often has fewer edges, and always has fewer parameters than $\mathcal{G}(V)$.

### 5.3 Complexity Analysis

We now give the complexity analysis of **EID**. We will consider a fixed query $(y, x, \mathfrak{Q}(V), \mathcal{G}(V))$, for which we let $V^* = An(Y)_{\mathcal{G}[V \setminus X]} \setminus X$. The original graph $\mathcal{G}(V)$ will be assumed to have $n$ nodes and $m$ edges, while the graph $\mathcal{G}[V^*]$ will be assumed to have $n^*$ nodes.

**Lemma 7** *Evaluating line 1 of **EID** has time complexity $O(n + m + |\mathfrak{Q}(V^* \mid x)|)$.*

**Lemma 8** *Evaluating line 2 of **EID** has time complexity $O(|\mathcal{I}(\mathcal{G})| \cdot n^2 \cdot (n + m))$.*

*Proof:* Finding ancestors of a set can be done using an $O(n)$ traversal, which is done possibly $O(n)$ times. Finding a smaller district containing the outcome is equivalent to the problem of updating connected sets in undirected graphs when edges are deleted one by one. An algorithm for this problem exists with time complexity $O(q + n \cdot m)$, where $q$ is the number of edge deletions [13]. This implies finding the intrinsic closure of any $\leftrightarrow$-connected $S$ can be done in time $O(n \cdot (n + m))$. It can be shown (we omit the proof for space reasons) that all elements in $\mathcal{I}(\mathcal{G})$ can be obtained from the set of closures of singleton nodes in $\mathcal{G}$ by successively taking unions and closures. This implies the overall complexity stated. □

Let $b = bw_\pi(\mathcal{G}[V^*], V^* \setminus Y)$, for some order $\pi$ which we assume **EID** is using.

**Lemma 9** *For each latent projection $\mathcal{G}_i$ obtained by eliminating the first $i$ nodes in $V^* \setminus Y$ in order $\pi$ from $\mathcal{G}[V^*]$, computing the set $\mathcal{I}(\mathcal{G}_i)$ has time complexity $O(|\mathcal{I}(\mathcal{G}_i)| \cdot n_i^2 \cdot (n_i + m_i))$, where $n_i$ is the number of nodes in $\mathcal{G}_i$ and $m_i$ is the number of edges in $\mathcal{G}_i$.*

**Lemma 10** *Line 3 of **sum-one** has time complexity $O(4^b)$.*

*Proof:* Given that we know the heads and tails of all parameters in $\mathcal{G}(V \setminus X)$, we can tell in $O(1)$ time whether a given parameter must be computed via line 3. The number of assignments $\nu$ is bounded by $2^b$, by definition. For each such assignment, we call **GMT** on each district in the partition of $I \cup \{X\}$, which is equivalent to calling **GMT** on the entire set. But this is exponential in the number of nodes assigned 1, and so is bounded by $2^b$. □

**Lemma 11** *Line 4 of **sum-one** has time complexity $O(n^* \cdot 2^b)$.*

*Proof:* For each element in $\mathfrak{Q}(V)$, we must perform a constant amount of work to determine whether the parameter stays unchanged or was computed via line 3. But $|\mathfrak{Q}(V)|$ has size at most $O(n^* \cdot 2^b)$. □

Noting that $n_i + m_i$ is bounded by $n + m$, and $|\mathcal{I}(\mathcal{G}_i)|$ is bounded by $n^* \cdot 2^b$, we put our results together to give the overall complexity of **EID**.

**Theorem 3** ***EID** has time complexity $O(n + m + |\mathfrak{Q}(V^* \mid x)| + (n^* - |Y|) \cdot (4^b + 2^b \cdot n^* \cdot n^2 \cdot (n + m)))$.*

We emphasize that our analysis of graph operations of **EID** is somewhat coarse, and the complexity of these operations may potentially be improved. We believe a more careful analysis would render the paper more complex for little return, since the main computational roadblock in practice is the exponential dependence on width. An even coarser grained expression for the complexity of **EID** is $O(poly(n) \cdot 4^b)$, where $poly(n)$ is some low order polynomial over $n$.

### 5.4 A Dynamic Programming Scheme for GMT

A Möbius transform [4] on a distribution over $n$ variables has time complexity $O(4^n)$. A dynamic programming algorithm exists which is capable of implementing the transform with time complexity $O(n \cdot 2^n)$. This algorithm is known as the Fast Möbius Transform [4]. We develop a similar scheme for our generalized Möbius transform. It is shown in Fig. 6.

**Theorem 4** *If every input to **FGMT** in the course of evaluating $P(X_V \mid X_W)$ is memoized (cached on evaluation), and assignments $\mu$ are processed in ascending order of cardinality of 1 assignments in $\mu$, then the* time and space complexity of using **FGMT** to evaluate $P(X_V \mid X_W)$ is $O(|V \cup W| \cdot 2^{|V \cup W|})$.

**Corollary 1** *If memoized **FGMT** is used instead of **GMT** in line 3b of **sum-one**, then the time complexity of **EID** is $O(n + m + |\mathfrak{Q}(V^* \mid x)| + (n^* - |Y|) \cdot (b \cdot 2^b + 2^b \cdot n^* \cdot n^2 \cdot (n + m)))$.*

A coarser expression for the overall complexity of **EID** is then $O(\text{poly}(n) \cdot b \cdot 2^b)$, where $\text{poly}(n)$ is a low order polynomial over $n$.

### 5.5 An Example

We illustrate the operation of the algorithm with an example. Consider the mixed graph in Fig. 7(a), parameterized by: $q_{x_1}$, $q_{x_2}(x_1)$, $q_{x_1,x_3}(x_2)$, $q_{x_3}(x_2)$, $q_{x_2,x_4}(x_1, x_3)$, $q_{x_4}(x_3)$, $q_{x_5,x_3,x_1}(x_2, x_4)$, $q_{x_5,x_3}(x_2, x_4)$, $q_{x_5}(x_4)$, for a total of 23 parameters. A more naive representation of the distribution $P(x_1, x_2, x_3, x_4, x_5)$ (including the parameterization in [8]) would contain 31 parameters. We wish to evaluate the query $P(x_5 \mid do(x_3))$. According to **EID**, $V^* = An(X_5)_{\mathcal{G}[X_1,X_2,X_4,X_5]} \setminus \{X_3\}$, which means the relevant parameter restriction computed in the first stage is $\mathfrak{Q}(X_5, X_4 \mid x_3)$, which is equal to $\{q_{x_5}(x_4), q_{x_4}(x_3)\}$. In the second stage, **EID** marginalizes out all nodes not in the outcome $X_5$ from this representation, which just entails marginalizing out $X_4$. This is done using **sum-one**, which uses **GMT** to compute $P(x_5 \mid do(x_4))$ and $P(x_4 \mid do(x_3))$, computes $P(x_5 \mid do(x_3)) = \sum_{x_4} P(x_5 \mid do(x_4)) P(x_4 \mid do(x_3))$, and converts back to a $q$ representation of $P(x_5 \mid do(x_3))$, which corresponds to the graph shown in Fig. 7(c). Note that using $q$ parameters allows us to compute precisely the effect expression in Section 2.3 without resorting to 32 conditional probabilities.

Note that the same running time will hold for evaluating the query $P(x_k \mid do(x_{k-2}))$ for any extension of the graph in Fig. 7(a) to $k$ nodes. This is in contrast to the original formulation of the **ID** algorithm, which will involve probability tables of size exponential in $k$.

## 6 Discussion

We have given an algorithm (**EID**) which can compute interventional marginal distributions in latent variable causal models. Our algorithm achieves some measure of computational efficiency by exploiting conditional and post-truncation independence constraints embedded in r-factorization. In the special case of no interventions, our algorithm can be viewed as an efficient inference scheme for graphical models of conditional independence containing latent variables. Given a maximum likelihood estimator for $q$ parameters of r-factorizing models, our algorithm can be viewed as giving an MLE for causal effects in such models.

Going beyond binary models, **EID** can be viewed as generalizing the g-formula for identifiable causal effects to mixed graph models, by noting that in such models, $P(y \mid do(x)) = \sum_{v^* \setminus y} P^*(V^*)$, where $V^*$ are observable ancestors of $Y$ in the graph after $do(x)$ was performed, and $P^*$ is a distribution obtained from $P(V)$ by "truncating out" all (r-)factors containing $X$.

We hasten to add that r-factorizing models are inherently conservative, in the sense that they make no assumptions about the latent variables. A statistical model which makes additional modeling assumptions on the latents is capable of more efficient parameterizations and algorithms. Consider a DAG model shown in Fig. 2(a), where nodes $\{X_1, X_2, \ldots\}$ are latent, and all nodes are binary. This model is parameterized by $P(x_1 = 0)$, $P(y_1 = 0 \mid x_1)$, and $P(y_i = 0 \mid x_{i-1}, x_i)$, $P(x_i = 0)$, for $i > 1$. Since all nodes are binary, the total number of parameters grows linearly with the total number of nodes in this model.

A latent projection of the DAG in Fig. 2(a) onto

---

function **FGMT**($\mathcal{G}, \mathfrak{Q}, \mu$)
INPUT: $\mathcal{G}$ a CADMG, $\mathfrak{Q}$ a set of $q_H(X_H = \mathbf{0} \mid x_T)$, for every head $H$ in $\mathcal{G}$, $\mu$ an assignment to $X_{V \cup W}$.
OUTPUT: The probability $p(X_V = \mu_V \mid X_W = \mu_W)$.
Let $\prec$ be a topological ordering on $V \cup W$.

0. If $\mathcal{G}$ is the empty graph, return 1.

1. If $\mathcal{G}$ has more than one district then call **FGMT**($\mathcal{G}[D], \mathfrak{Q}(D \mid \mu_{Pa(D) \setminus D}), \mu_{D \cup Pa(D)})$ on each district $D \in \mathcal{D}(\mathcal{G})$; return the product of the results.

2. If $\mathcal{D}(\mathcal{G}) = \{D\}$ then find $H = rh(D)$;

   (a) If $\mu_H = \mathbf{0}$ then
      (i) $q^*_{(D \setminus H) \cup W} \leftarrow$ **FGMT**($\mathcal{G}_{(D \setminus H) \cup W}$, $\mathfrak{Q}(D \setminus H \mid \mu_{Pa(D) \setminus D}), \mu_{(D \setminus H) \cup W})$.
      (ii) return $q^*_{(D \setminus H) \cup W} \times q_H(\mu_{tail(H)})$.

   (b) Else find the first $Y$ under $\prec$ such that $Y \in H$ and $\mu_Y = 1$;
      (i) $q^*_{(D \setminus \{Y\}) \cup W} \leftarrow$ **FGMT**($\mathcal{G}_{(D \setminus \{Y\}) \cup W}$, $\mathfrak{Q}(D \setminus \{Y\} \mid \mu_{Pa(D) \setminus D}), \mu_{(D \setminus \{Y\}) \cup W})$.
      (ii) $q^0_{D \cup W} \leftarrow$ **FGMT**($\mathcal{G}_{D \cup W}, \mathfrak{Q}, (\mu_{(D \setminus \{Y\}) \cup W}, y = 0))$.
      (iii) return $q^*_{(D \setminus \{Y\}) \cup W} - q^0_{D \cup W}$.

Figure 6: Fast Generalized Möbius Transform. Each input is memoized (cached).

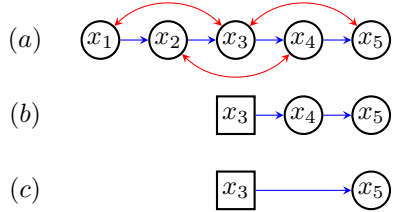

Figure 7: (a) A graph where we wish to evaluate $P(x_5 \mid do(x_3))$. (b) A graph $\mathcal{G}[An(X_5)_{\mathcal{G}[X_1,X_2,X_4,X_5]} \setminus \{X_3\}]$. (c) A graph representing the output distribution $P(x_5 \mid do(x_3))$.

$\{Y_1, Y_2, \ldots\}$ is shown in Fig. 2(b). It is not difficult to show that the number of $q$ parameters in graphs of the type shown in Fig. 2(b) grows as $O(k^2)$ for a graph with $k$ nodes. The DAG model with latents in Fig. 2(a) posits additional assumptions about the latent variables beyond those implied by the conditional independence structure of its graph, namely that the number of states of the latent variables $X_i$ does not depend on the total number of such variables. On the other hand, if we are not willing to commit to any assumptions about latent variables, and only wish to assume conditional and post-truncation independences implied by a particular mixed graph, we believe using r-factorization is in some sense "optimal."

Our algorithm can be used to obtain identifiable conditional interventional distributions of the form $P(y \mid z, do(x))$. A known result [14] implies identification problems for such distributions are always equivalent to identification problems for an unconditional effect of the form $P(y' \mid do(x'))$, where $P(y \mid z, do(x)) = P(y' \mid do(x'))/P(z' \mid do(x'))$, for some $Z' \subseteq Y'$. This result means, we can use our algorithm to obtain $q$ parameters for $P(y' \mid do(x'))$, and obtain our answer by one final conditioning operation.

### Acknowledgments

This research was supported by the U.S. National Science Foundation (CRI 0855230) and U.S. National Institutes of Health (R01 AI032475).